\documentclass[letterpaper, 10 pt, conference]{ieeeconf}  

\IEEEoverridecommandlockouts                              

\usepackage[caption=false]{subfig}
\usepackage{amsmath} 

\usepackage[Algorithm]{algorithm}
\usepackage{algorithmic}
\usepackage{booktabs}
\usepackage{todonotes}
\usepackage{pgfplots}

\makeatletter
\let\NAT@parse\undefined
\makeatother
\usepackage[colorlinks,citecolor=red,urlcolor=blue,bookmarks=false,hypertexnames=true]{hyperref}
\usepackage[capitalise]{cleveref}

\definecolor{basiccolor}{HTML}{1b9e77}
\definecolor{densecolor}{HTML}{d95f02}
\definecolor{sparsecolor}{HTML}{7570b3}
\definecolor{differencecolor}{HTML}{e7298a}
\definecolor{othercolor}{HTML}{66a61e}

\title{\LARGE \bf
	Real-time event simulation with frame-based cameras
}

\author{Andreas Ziegler, Daniel Teigland, Jonas Tebbe, Thomas Gossard and Andreas Zell
	\thanks{The authors are with the Cognitive Systems Group, Dept. Informatics, University of Tuebingen.
		Corresponding author {\tt\small andreas.ziegler@uni-tuebingen.de}.\newline
		This research was funded by Sony AI.}%
}

\usepackage{url}

\begin{document}
	
	\maketitle
	\thispagestyle{empty}
	\pagestyle{empty}

	\begin{abstract}
		Event cameras are becoming increasingly popular in robotics and computer vision due to their beneficial properties, e.g., high temporal resolution, high bandwidth, almost no motion blur, and low power consumption.
		However, these cameras remain expensive and scarce in the market, making them inaccessible to the majority.
		Using event simulators minimizes the need for real event cameras to develop novel algorithms.
		However, due to the computational complexity of the simulation, the event streams of existing simulators cannot be generated in real-time but rather have to be pre-calculated from existing video sequences or pre-rendered and then simulated from a virtual 3D scene.
		Although these offline generated event streams can be used as training data for learning tasks, all response time dependent applications cannot benefit from these simulators yet, as they still require an actual event camera.
		This work proposes simulation methods that improve the performance of event simulation by two orders of magnitude (making them real-time capable) while remaining competitive in the quality assessment.
	\end{abstract}
	
	\section*{SUPPLEMENTARY MATERIAL}
	The project’s code and additional resources are available at \url{https://cogsys-tuebingen.github.io/realtime_event_simulator/}

	\section{INTRODUCTION}
	
	In the field of robotics and computer vision, event cameras have been rapidly gaining in popularity.
	Event cameras work fundamentally differently from traditional frame-based cameras. 
	Instead of capturing an entire frame of pixels at once, an event camera registers log-scale changes in brightness on a pixel level.
	If the change is large enough, the camera emits an asynchronous event.
	The advantages of event cameras are that they can register events just microseconds apart, resulting in lower latency and higher temporal resolution ($\mu s$ instead of $ms$) compared to frame-based cameras.
	These cameras also come with other great potential benefits, such as a high dynamic range and low power consumption~\cite{Gallego20pami}.
	This technology allows for super low latency applications that, when implemented correctly, can leverage all this data to outperform the best traditional computer vision algorithms in various robotics and computer vision tasks.
	Some useful applications include self-driving cars~\cite{Maqueda18cvpr}, general purpose object tracking~\cite{Perot20neurips}, image segmentation~\cite{Alonso19cvprw, Zhou21tnnls}, high frame rate video reconstruction~\cite{Pan20pami}, accurate depth perception~\cite{Muglikar213dv} and advanced frame interpolation with optical flow reconstruction~\cite{Tulyakov21cvpr} and robotics applications like control tasks~\cite{Dimitrova20icra},~\cite{Vitale21icra}, and odometry~\cite{Hidalgo22cvpr}.
	For some applications, e.g., self-driving cars~\cite{Maqueda18cvpr}, it is essential that this data is processed in real-time.
	
	Although the first event camera was released more than a decade ago, event cameras remain expensive ($\sim6000$\texteuro) and scarce in the market, making them inaccessible to many researchers, companies, and hobbyists alike.
	This work proposes using frame-based cameras to simulate this event data.
	Traditional frame-based cameras remain cheap and readily available, making this approach an interesting alternative. 
	
	Over the last few years, several methods have been published that attempt to simulate event data.
	While these recent publications produce great results of high quality in the event simulation, the computation required for just one frame can take up to multiple seconds, eliminating the possibility of real-time use.
	To truly use a frame-based camera as a replacement for an event camera in robotics applications, these simulators would have to run at much higher speeds. \par
	This work specifically attempts to address this problem by presenting methods that aim to drastically improve the runtime of these simulations while remaining competitive in quality, to be of use for real-time robotics applications.
	
	\textbf{Contributions} of this work are as follows:
	\begin{itemize}
		\item Optical flow based event simulation methods running in real-time  
		\item A novel event simulation method which leverages the sparsity of events in the interpolation to further improve the runtime
		\item Qualitative and quantitative results comparing our event simulation methods with existing ones and real event cameras, and a guideline, when to use which simulator
	\end{itemize}
	The rest of the paper is organized as follows: \Cref{sec:related_work} presents the related work.
	In \cref{sec:method} we present our methods.
	In \cref{sec:evaluation} we present qualitative and quantitative comparisons.
	Conclusions are drawn in \cref{sec:conculsion}.
	
	\section{RELATED WORK}\label{sec:related_work}
	
	One of the first event simulators was described in~\cite{Kaiser16simpar}.
	This event simulator subtracts consecutive frames and afterward applies a threshold to retain only pixels that have a significant difference.
	The approach in~\cite{Nehvi21cvprw} is also based on subtracting frames, however, their contribution is the differentiability of the event simulator, which is needed for backpropagation.
	While our approach subtracts consecutive frames, we use optical flow to interpolate the given input frames, leading to a temporal resolution closer to one of the real event cameras.
	Approaches such as~\cite{Mueggler17ijrr},~\cite{Wenbin18bmvc}, and~\cite{Radomski21arxiv} use a rendering engine to capture images from a 3D scene.
	
	In~\cite{Rebecq18corl}, the authors developed an event simulator tightly coupled with the rendering engine to adaptively sample the visual signal from the virtual camera sensor along a given camera trajectory.
	This allows generating simulated events asynchronously, whereas previous simulators only provide synchronous events.
	This approach also simulates some linear noise but neglects non-linear noise.
	In~\cite{Gehrig20cvpr}, the authors extended their previous work~\cite{Rebecq18corl} by combining it with a frame interpolation method to generate events from videos.
	Similar to~\cite{Gehrig20cvpr},~\cite{Hu21cvprw} converts regular video sequences to events.
	Compared to previous work,~\cite{Hu21cvprw} includes more realistic noise models and therefore comes closer to the output of real event cameras.
	In~\cite{Joubert21fnins}, the authors go one step further and add several improvements and additions.
	The model notably improves the noise simulation by directly mapping noise distributions from real pixels. 
	In another recent approach, the authors developed a GAN-style neural network to regress an event volume (an event representation)~\cite{Zhu19arxiv}.
	To train the network, two consecutive frames and the corresponding events were given.
	On inference time, only two consecutive frames are needed.
	While the approach can approximate the true noise distributions of real events more accurately, it does not seem to run in real-time, and the output is an event representation and not the events themselves.
	
	Most of the existing event simulators focus on the realistic event output of the simulator.
	This comes at the cost of a high computational load, and therefore an increased runtime.
	For many applications, especially learning-based solutions, such event simulators are ideal.
	However, many robotics applications favor real-time capability over realistic events with accurate noise.
	Our event simulators have a less realistic noise model, but do run in real-time while remaining competitive in quality.
	To the best of our knowledge, this work presents the first method to simulate events given the output of a frame-based camera in real-time.
	
	\section{METHOD}\label{sec:method}
	
	We present three different methods to simulate events.
	All of them are based on the subtraction of consecutive frames, as done in~\cite{Kaiser16simpar}.
	We call the result of the subtraction the \textit{difference frame}.
	In contrast to event cameras, we do not use the log difference but the linear difference.
	The quality improvement of log differences is small and comes with a high computational load (delay in ms depending on the interpolated frames).
	Further, this work does not address the noise and artifacts present in data from a real event camera.

	The temporal resolution of frame-based cameras is much lower than that of event cameras~\cite{Gallego20pami}.
	To counteract this, we use interpolation to artificially increase the frame rate.
	Interpolation is a common step in event simulation and is also used in~\cite{Gehrig20cvpr} and~\cite{Hu21cvprw}.
	For interpolation, we use optical flow algorithms and a simple but fast linear interpolation approach.
	We used the fastest optical flow algorithms evaluated in~\cite{Kroeger16eccv}.
	
	Next follows the detailed description of the three methods.
	In \cref{subsec:dense_interpolation}, we describe a method based on dense frame interpolation.
	\Cref{subsec:sparse_interpolation} introduces a method that uses sparse frame interpolation, and the technique described in \cref{subsec:difference_interpolation} interpolates only the \textit{difference frames}.
	
	\subsection{DENSE FRAME INTERPOLATION}\label{subsec:dense_interpolation}
	
	This method calculates the dense optical flow between two consecutive camera frames in the first step.
	This dense optical flow is then used to interpolate between the two consecutive camera frames.
	Given these interpolated frames, consecutive frames are subtracted to get \textit{difference frames}.
	In the last step, the \textit{difference frames} are thresholded (with $C_{\text{pos}}$ and $C_{\text{neg}}$) to determine all the pixels that should emit an event, and the corresponding polarity of the event according to
	\begin{align*}
		\operatorname{e}_{\text{pos}}(x, y) = \begin{cases}\text{pos. event}, &\text{if} \operatorname{difference\_frame}(x, y)>C_{\text{pos}} \\ \text{no event}, &\text {otherwise}\end{cases} \\
		\operatorname{e}_{\text{neg}}(x, y) = \begin{cases}\text{neg. event}, &\text{if} \operatorname{difference\_frame}(x, y)<C_{\text{neg}} \\ \text{no event}, &\text {otherwise}.\end{cases}
	\end{align*}
	
\begin{figure}[thpb]
		\centering
		\includegraphics[width=0.7\linewidth]{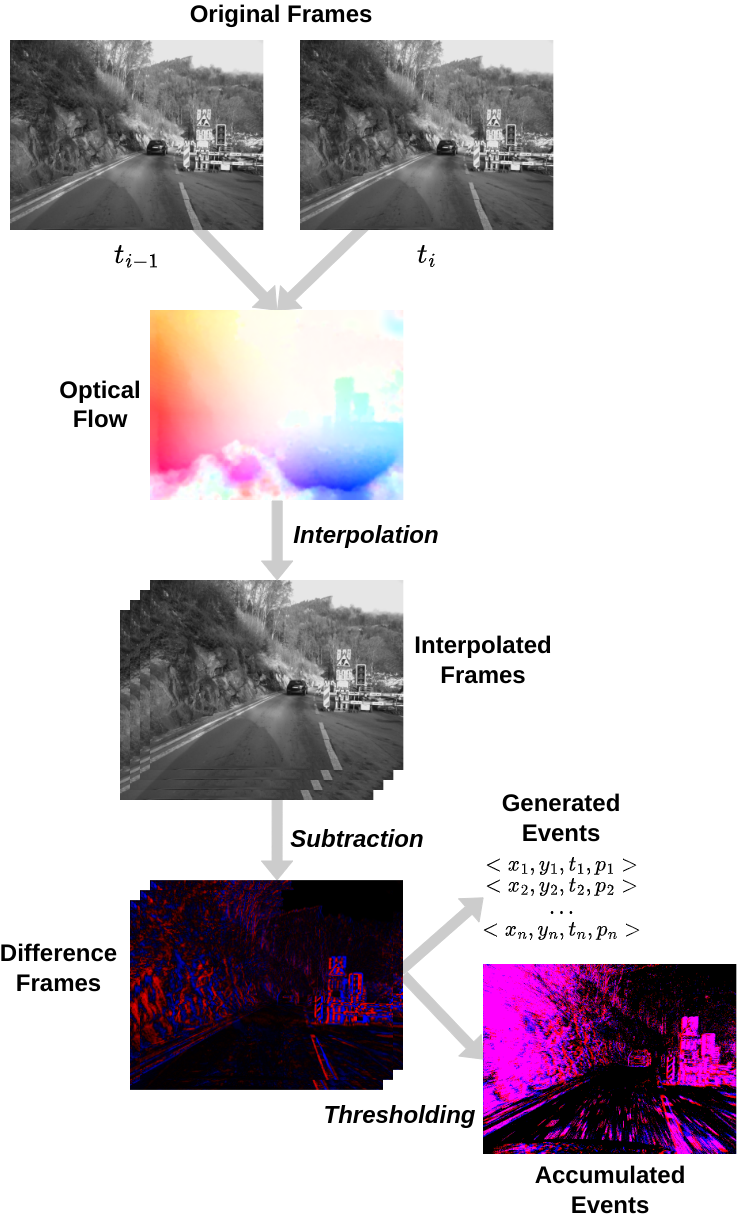}
		\caption{
			The process of the dense frame interpolation method, given the consecutive camera frames at $t_{i-1}$ and $t_{i}$ as input.
			The output is either the events in tuples of the form $<$$x, y, t, p$$>$ with the location of the event $(x, y)$, the time stamp $t$ and the polarity of the event $p$ or a frame with the accumulated events.
			\textit{Processing steps are written with italic letters} whereas (intermediate) outputs are written with normal letters.
			For more details we refer to \cref{subsec:dense_interpolation}.
		}
		\label{fig:dense_method}
	\end{figure}

	The process of this approach is illustrated in \cref{fig:dense_method}.
	
	The output is either the events in tuples of the form $<$$x, y, t, p$$>$ with the location of the event $(x, y)$, the time stamp $t$ and the polarity of the event $p$ or a frame with the accumulated events.
	
	In order to use this algorithm in real-time applications, the choice of the dense optical flow algorithm is very important, as its speed is the main bottleneck for the performance of this method.	
	
	\subsection{SPARSE FRAME INTERPOLATION}\label{subsec:sparse_interpolation}
	
	This method is similar to the dense frame interpolation approach described in \cref{subsec:dense_interpolation} with the difference that sparse optical flow is used to interpolate between two consecutive camera frames.
	
	The motivation to use sparse optical flow is to increase the performance by reducing the number of calculations for the frame interpolation step.
	Dense optical flow algorithms consider the entire frame while only a fraction of the pixels in a frame is actually changing a lot, most of the time.
	Real event cameras use a threshold to detect when an event at a particular pixel has occurred.
	Analogously to an event camera, all pixels that have not changed within a given threshold from one frame to the next can be disregarded for the optical flow estimation.
	This way, the computation can be sped up as long as the number of pixels that have changed is substantially lower than the total number of pixels in the frame.
	The selected pixels can be estimated with any sparse optical flow algorithm.
	
	It is important to note that sparse optical flow estimations are usually slower than dense algorithms per pixel, so this algorithm benefits from static backgrounds containing small moving subjects.
	
	\subsection{DIFFERENCE FRAME INTERPOLATION}\label{subsec:difference_interpolation}
	
	Interpolating regular frames is a complex problem with many unforeseen edge cases such as, e.g., occlusions, and is computationally expensive.
	Traditional interpolation algorithms all rely on the basic assumption that the area of pixels that make up an object will not change drastically in color, brightness, or size from one frame to another.
	This assumption breaks down in scenes where occlusions occur.
	Traditional optical flow algorithms cannot find the original pixels, and artifacts appear in the optical flow estimation.
	
	To reduce the possibility of such edge cases and to further decrease the computational load of the interpolation step, this approach only interpolates between consecutive \textit{difference frames} rather than between consecutive camera frames. 
	This method reduces complexity by only interpolating between the events themselves since we assume that the \textit{difference frames} represent the events.
	
\begin{figure}[h!]
		\centering
		\includegraphics[width=0.8\linewidth]{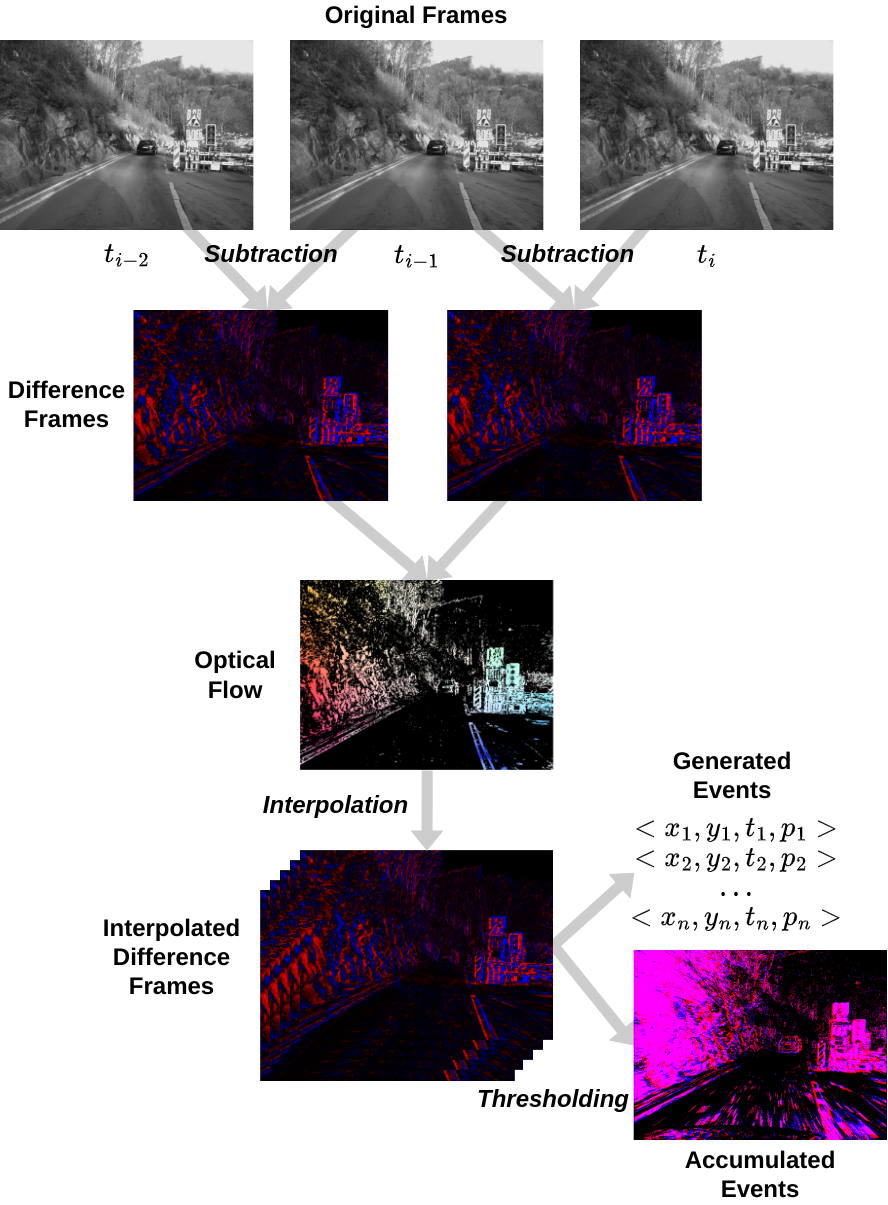}
		\caption{
			The process of the difference frame interpolation method, given the consecutive camera frames at $t_{i-2}$, $t_{i-1}$ and $t_{i}$ as input.
			The output is either the events in tuples of the form $<$$x, y, t, p$$>$ with the location of the event $(x, y)$, the time stamp $t$ and the polarity of the event $p$ or a frame with the accumulated events.
			\textit{Processing steps are written with italic letters} whereas (intermediate) outputs are written with standard letters.
			For more details we refer to \cref{subsec:difference_interpolation}.
		}
		\label{fig:difference_method}
	\end{figure}

	Given three consecutive camera frames, two \textit{difference frames} are calculated in the first step.
	The optical flow between the two \textit{difference frames} is calculated in the next step.
	With this optical flow, additional \textit{difference frames} can be interpolated between the two \textit{difference frames}, from the first step.
	The process of this approach is illustrated in \cref{fig:difference_method}.
	
	This method makes some strict assumptions about the original frame sequence.
	The objects' movement that generates events must be of constant linear velocity.
	Furthermore, the perspective of these objects in the video cannot change over time.
	Given these assumptions, this method is specially targeted for tasks like, e.g., object tracking.
	
	\section{EVALUATION}\label{sec:evaluation}
	
	We evaluate our proposed simulation methods in three different ways.
	We start in \cref{subsec:speed} with a comparison of the runtimes of the different event simulation methods.
	In \cref{subsec:qualitative_comparison}, we qualitatively compare the output of our methods to real events and the output of another event simulator.
	Next to the qualitative comparison, we compare some statistics of the simulated, the real events, and two other event simulators in \cref{subsec:statistical_comparison}.
	
	\subsection{RUNTIMES}\label{subsec:speed}
	
\begin{figure}[h!]
		\centering
		\begin{tikzpicture} 
			\begin{axis}[ 
				width=0.53\textwidth,
				height=6.6cm,
				title style={align=center,text width=7.0cm},
				title={RUNTIME COMPARISON [MS] ON HIGH EVENTS-PER-FRAME DATA},
				ymode=log,
				xtick=data,
				xtick pos=bottom,
				ticklabel style={font=\footnotesize},
				ybar, 
				ymin=0.3,
				x tick label style={rotate=70,align=right,anchor=east,text width=3.0cm},
				every axis plot/.append style={ybar,bar shift=0pt},
				xtick={1,2,3,4,5,6,7,8,9,10,11},
				x tick style={draw=none},
				xticklabels={Difference frame,
					Dense int. w. DIS~\cite{Kroeger16eccv} on lq. preset,
					Dense int. w. DIS~\cite{Kroeger16eccv} on hq. preset,
					Dense int. w. Farneback~\cite{Farnebck03scia},
					Sparse int. w. LK~\cite{Kanade81iuw},
					Difference int. w. LK~\cite{Kanade81iuw},
					Dense int. w. Farneback~\cite{Farnebck03scia} (GPU),
					Sparse int. w. LK~\cite{Kanade81iuw} (GPU),
					Difference int. w. LK~\cite{Kanade81iuw} (GPU),
					v2e~\cite{Hu21cvprw},
					vid2e~\cite{Gehrig20cvpr}*},
				extra y ticks = 30,
				extra y tick labels={},
				extra y tick style={grid=major,major grid style={thick,draw=blue}}
				]
				\addplot+[basiccolor,fill=basiccolor!50,error bars/.cd, y dir=both,y explicit] coordinates { 
					(1,0.43)};
				\addplot+[densecolor,fill=densecolor!50,error bars/.cd, y dir=both,y explicit] coordinates { 
					(2,17.69) +- (0.0, 0.15)};
				\addplot+[densecolor,fill=densecolor!50,error bars/.cd, y dir=both,y explicit] coordinates {
					(3,58.30) +- (0.0, 0.15)};
				\addplot+[densecolor,fill=densecolor!50,error bars/.cd, y dir=both,y explicit] coordinates {
					(4,72.81) +- (0.0, 0.17)};
				\addplot+[sparsecolor,fill=sparsecolor!50,error bars/.cd, y dir=both,y explicit] coordinates {
					(5,144.99) +- (0.0, 0.27)};
				\addplot+[differencecolor,fill=differencecolor!50,error bars/.cd, y dir=both,y explicit] coordinates {
					(6,67.64) +- (0.0, 0.3)};
				\addplot+[densecolor,fill=densecolor!50,error bars/.cd, y dir=both,y explicit] coordinates {
					(7,20.22) +- (0.0, 0.20)};
				\addplot+[sparsecolor,fill=sparsecolor!50,error bars/.cd, y dir=both,y explicit] coordinates {
					(8,22.45) +- (0.0, 0.19)};
				\addplot+[differencecolor,fill=differencecolor!50,error bars/.cd, y dir=both,y explicit] coordinates {
					(9,10.07) +- (0.0, 0.06)};
				\addplot+[othercolor,fill=othercolor!50,error bars/.cd, y dir=both,y explicit] coordinates {
					(10,71741.58) +- (0.0, 622.14)};
				\addplot+[othercolor,fill=othercolor!50,error bars/.cd, y dir=both,y explicit] coordinates {
					(11,1420506.78) +- (0.0, 127186.50)};
				\draw[thick,black,dotted](55.0, -2.0) -- (55.0, 20.0);
				\node[black] at (20,9) {\large CPU};
				\node[black] at (70,9) {\large GPU};
			\end{axis} 
		\end{tikzpicture}
		\vspace{-1cm}
		\caption{Runtime comparison [ms] on high events-per-frame footage (car in Zurich~\cite{Gehrig21ral}) with a resolution of $640 \times 480$ pixels.
			LK stands for Lucas Kanade~\cite{Kanade81iuw}.
			*For vid2e we only measured the time for the event generation but not for the upsampling due to the time-consuming upsampling.
			The bars visualize the mean value and the standard deviation.
			The {\color{blue}blue line} indicates a runtime of 30ms which we consider as real-time capable in this work.}
		\label{fig:zurich_times}
	\end{figure}
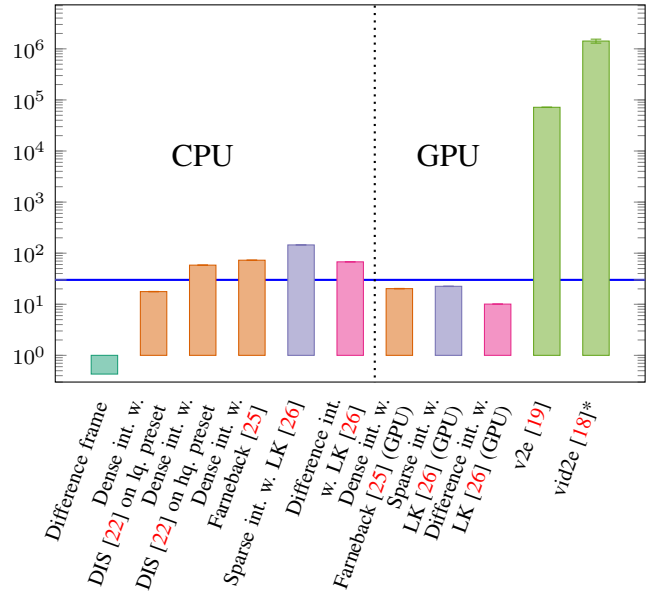

	We compare the runtimes of our proposed event simulation methods in two different scenarios.
	The first scenario is a dynamic scene with a high events-per-frame ratio.
	This footage is taken from~\cite{Gehrig21ral}.
	The dataset contains recorded events and camera frames at $20$ fps from a car driving around.
	We took a frame sequence with $537$ frames from the {\it interlaken\_00\_c} sequence.
	The frames with an original resolution of $1440 \times 1080$ pixels were resized to the resolution of the event camera ($640 \times 480$ pixels) before passing it to the event simulators.
	For this dataset, we used $20$ interpolated (difference) frames.
	The results of this experiment are shown in \cref{fig:zurich_times}.
	
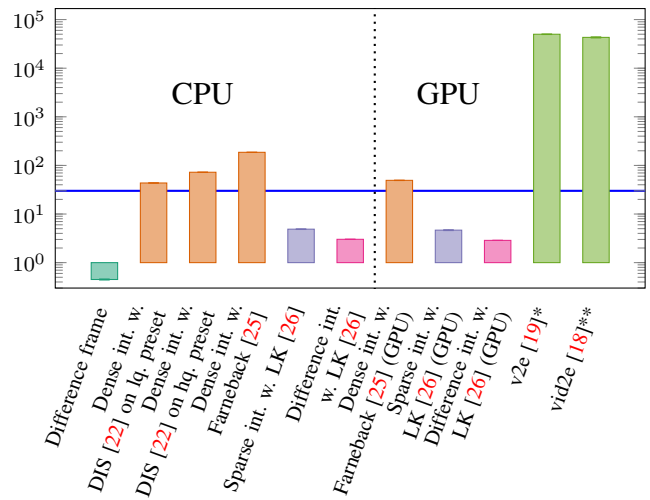
\begin{figure}[h!]
		\centering
		\begin{tikzpicture} 
			\begin{axis}[ 
				width=0.53\textwidth,
				height=5.3cm,
				title style={align=center,text width=7.0cm},
				title={RUNTIME COMPARISON [MS] ON LOW EVENTS-PER-FRAME DATA},
				ymode=log,
				xtick=data,
				xtick pos=bottom,
				ticklabel style={font=\footnotesize},
				ybar, 
				ymin=0.3,
				x tick label style={rotate=70,align=right,anchor=east,text width=3.0cm},
				every axis plot/.append style={ybar,bar shift=0pt},
				xtick={1,2,3,4,5,6,7,8,9,10,11},
				x tick style={draw=none},
				xticklabels={Difference frame,
					Dense int. w. DIS~\cite{Kroeger16eccv} on lq. preset,
					Dense int. w. DIS~\cite{Kroeger16eccv} on hq. preset,
					Dense int. w. Farneback~\cite{Farnebck03scia},
					Sparse int. w. LK~\cite{Kanade81iuw},
					Difference int. w. LK~\cite{Kanade81iuw},
					Dense int. w. Farneback~\cite{Farnebck03scia} (GPU),
					Sparse int. w. LK~\cite{Kanade81iuw} (GPU),
					Difference int. w. LK~\cite{Kanade81iuw} (GPU),
					v2e~\cite{Hu21cvprw}*,
					vid2e~\cite{Gehrig20cvpr}**},
				extra y ticks = 30,
				extra y tick labels={},
				extra y tick style={grid=major,major grid style={thick,draw=blue}},
				] 
				\addplot+[basiccolor,fill=basiccolor!50,error bars/.cd, y dir=both,y explicit] coordinates {
					(1,0.45) +- (0.0, 0.01)};
				\addplot+[densecolor,fill=densecolor!50,error bars/.cd, y dir=both,y explicit] coordinates {
					(2,43.51) +- (0.0, 0.32)};
				\addplot+[densecolor,fill=densecolor!50,error bars/.cd, y dir=both,y explicit] coordinates {
					(3,72.37) +- (0.0, 0.26)};
				\addplot+[densecolor,fill=densecolor!50,error bars/.cd, y dir=both,y explicit] coordinates {
					(4,186.06) +- (0.0, 0.55)};
				\addplot+[sparsecolor,fill=sparsecolor!50,error bars/.cd, y dir=both,y explicit] coordinates {
					(5,4.88) +- (0.0, 0.07)};
				\addplot+[differencecolor,fill=differencecolor!50,error bars/.cd, y dir=both,y explicit] coordinates {
					(6,3.02) +- (0.0, 0.01)};
				\addplot+[densecolor,fill=densecolor!50,error bars/.cd, y dir=both,y explicit] coordinates {
					(7,49.30) +- (0.0, 0.42)};
				\addplot+[sparsecolor,fill=sparsecolor!50,error bars/.cd, y dir=both,y explicit] coordinates {
					(8,4.66) +- (0.0, 0.09)};
				\addplot+[differencecolor,fill=differencecolor!50,error bars/.cd, y dir=both,y explicit] coordinates {
					(9,2.87) +- (0.0, 0.01)};
				\addplot+[othercolor,fill=othercolor!50,error bars/.cd, y dir=both,y explicit] coordinates {
					(10,49934.16) +- (0.0, 618.19)};
				\addplot+[othercolor,fill=othercolor!50,error bars/.cd, y dir=both,y explicit] coordinates {
					(11,43040.42) +- (0.0, 1047.53)};
				\draw[thick,black,dotted](55.0, -2.0) -- (55.0, 15.0);
				\node[black] at (20,8) {\large CPU};
				\node[black] at (70,8) {\large GPU};
			\end{axis} 
		\end{tikzpicture}
		\vspace{-1cm}
		\caption{Runtime comparison [ms] on low events-per-frame footage (table-tennis ball) with a resolution of $1280 \times 720$ pixels.
			LK stands for Lucas Kanade~\cite{Kanade81iuw}.
			*Due to memory constraints on the GPU, we had to limit the output for v2e to $640 \times 420$ pixels.
			**For vid2e we only measured the time for the event generation but not for the upsampling due to the time-consuming upsampling.
			The bars visualize the mean value and the standard deviation.
			The {\color{blue}blue line} indicates a runtime of 30ms, which we consider as real-time capable in this work.}
		\label{fig:ball_times}
	\end{figure}

	We use a less dynamic scene with a low events-per-frame ratio in the second scenario.
	Specifically, we recorded a flying table tennis ball in front of a static background with a FLIR CHAMELEON 3 camera at $150$ fps and a Prophesee Gen4 event camera~\cite{Finateu20isscc}.
	This frame sequence contains $158$ frames.
	We also resized the input frames for this dataset from $1280 \times 1024$ pixels (the camera's resolution) to $1280 \times 720$ pixels (the resolution of the event camera), before passing it to the event simulators.
	For this dataset, we used $10$ interpolated (difference) frames.
	The results of this experiment are shown in \cref{fig:ball_times}.
	
	Next to the runtimes of our three proposed methods, we also list the runtimes for calculating the \textit{difference frame} as a reference.
	We also compare our event simulation methods to vid2e~\cite{Gehrig20cvpr} and v2e~\cite{Hu21cvprw}, two state-of-the-art event simulators.
	
	We used different optical flow algorithms for all three event simulation methods since the simulation methods heavily depend on the used optical flow implementation.
	
	We chose $C_{\text{pos}}$, $C_{\text{neg }}$, and the number of interpolated frames so that the generated events resemble the ground truth accumulated event frame.
	These values are listed in \cref{tab:evaluation_settings}.
	
	\begin{table}[h!]
		\centering
		\caption{$C_{\text{pos}}$, $C_{\text{neg }}$ values and the number of interpolated frames}
		\begin{tabular}{l|l|l|l}
		\textbf{Method} & \textbf{$C_{\text{pos}}$} & \textbf{$C_{\text{neg}}$} & \textbf{\# int. frames} \\
		\hline
		\hline
		{\color{densecolor} Dense interpolation} & 2 & -2 & 10 \\
		\hline
		{\color{sparsecolor} Sparse interpolation} & 10 & -10 & 10 \\
		\hline
		{\color{differencecolor} Difference interpolation} & 20 & -20 & 10 \\
		\hline
		{\color{othercolor} v2e}~\cite{Hu21cvprw} & 0.3 & -0.3 & ? \\
		\hline
		{\color{othercolor} vid2e}~\cite{Gehrig20cvpr} & 0.2 & -0.2 & ? \\
		\end{tabular}
		\label{tab:evaluation_settings}
	\end{table}
	
	All the experiments were run on a desktop with an Intel i7 9700 3.0 GHz CPU, 32GB RAM, and an NVIDIA GeForce RTX 2080 Ti.
	For each algorithm, the video sequence was loaded into RAM before the time was measured.
	Then the entire video is passed to each algorithm frame by frame.
	After the last frame, the time measurement was stopped.
	We recorded the runtimes of 10 runs.
	From these, the mean runtime and the standard deviation were calculated.
	
	\begin{figure*}[h!]
		\centering
		\subfloat[Grayscale frame (for reference)\label{subfig:grayscale_car}]{%
			\includegraphics[width=.15\textwidth]{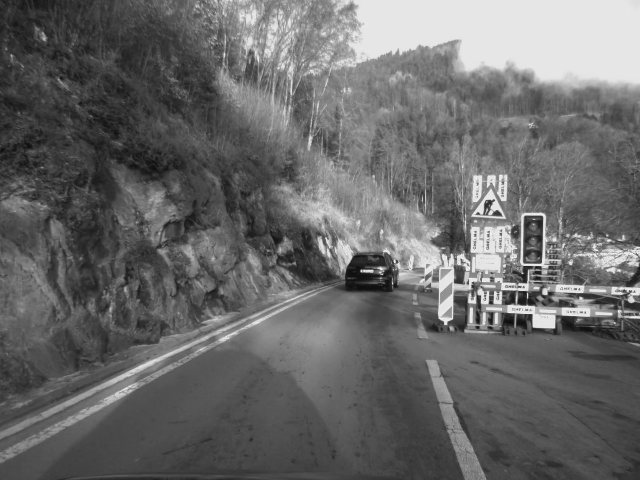}}
		\hfill
		\subfloat[Ground truth accumulated events frame\label{subfig:accum_events_car}]{%
			\includegraphics[width=.15\textwidth]{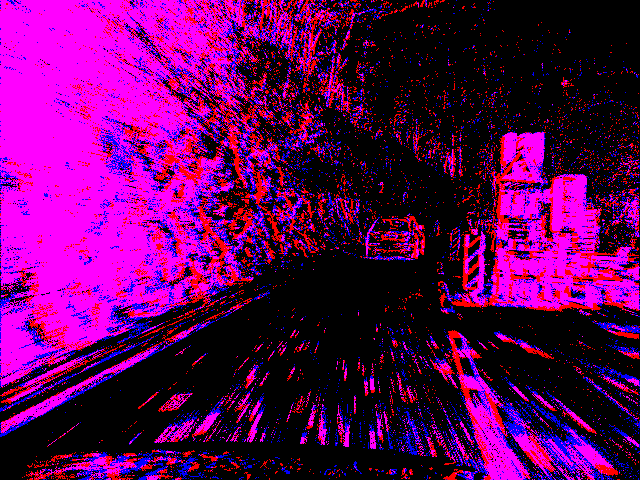}}
		\hfill
		\subfloat[\color{basiccolor}{Difference frame}\label{subfig:difference_frame_car}]{%
			\includegraphics[width=.15\textwidth]{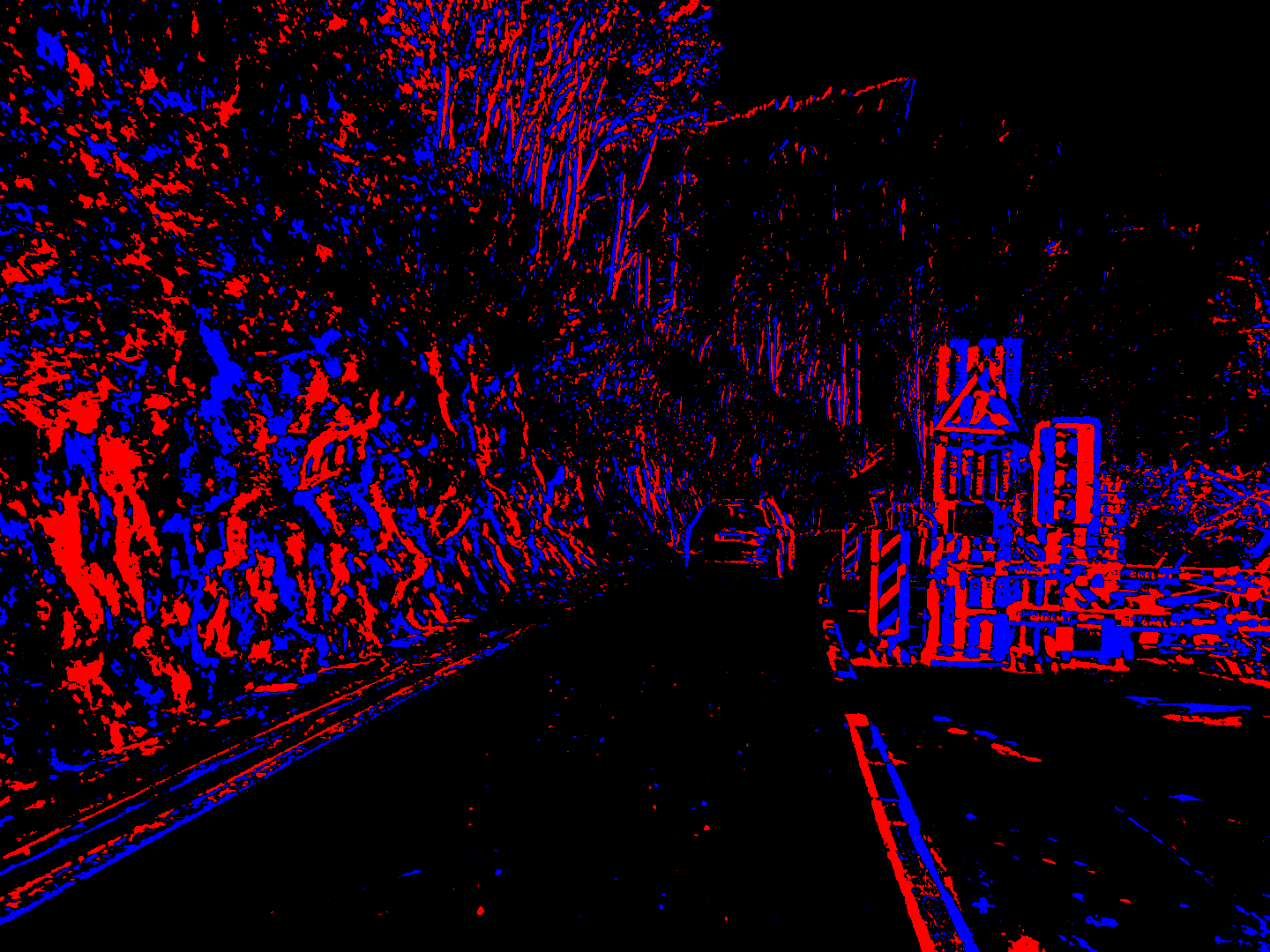}}
		\hfill
		\subfloat[{\color{densecolor}{Dense int. w. DIS}}~\cite{Kroeger16eccv} on lq. preset\label{subfig:dense_dis_low_car}]{%
			\includegraphics[width=.15\textwidth]{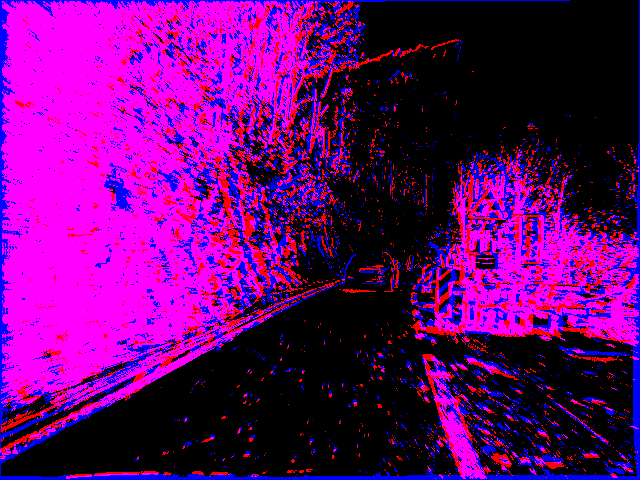}}
		\hfill
		\subfloat[{\color{densecolor}{Dense int. w. DIS}}~\cite{Kroeger16eccv} on hq. preset\label{subfig:dense_dis_high_car}]{%
			\includegraphics[width=.15\textwidth]{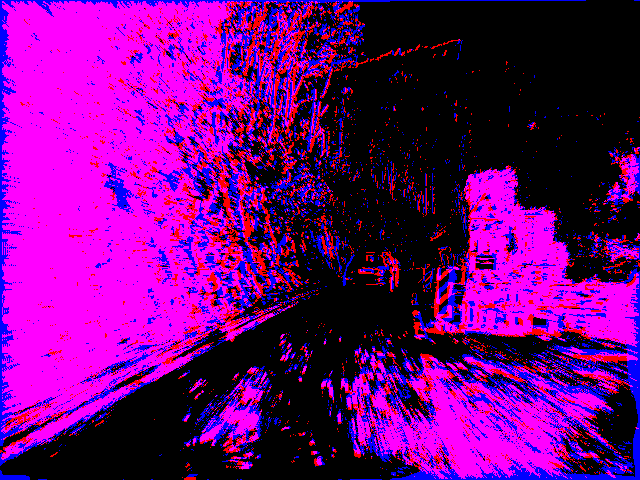}}
		\hfill
		\subfloat[{\color{densecolor}{Dense int. w. Farneback}}~\cite{Farnebck03scia} (CPU)\label{subfig:dense_farneback_cpu_car}]{
			\includegraphics[width=.15\textwidth]{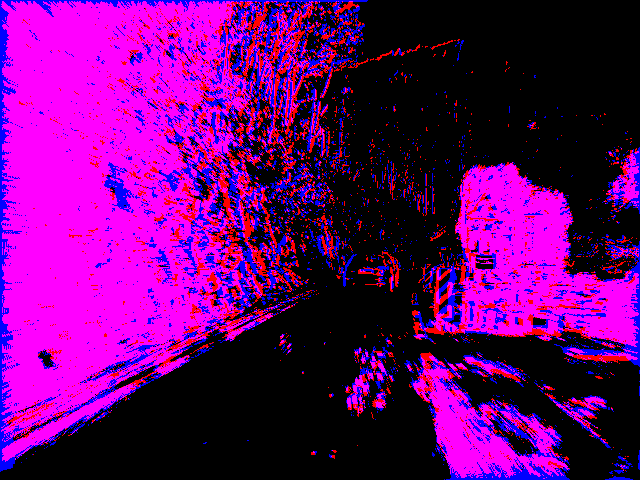}}

		\subfloat[{\color{sparsecolor}{Sparse int. w. LK}}~\cite{Kanade81iuw} (CPU)\label{subfig:sparse_lukas_cpu_car}]{%
			\includegraphics[width=.15\textwidth]{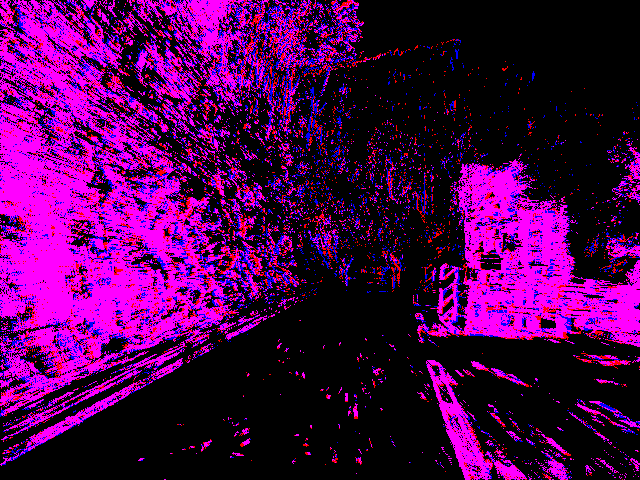}}
		\hfill
		\subfloat[{\color{differencecolor}{Difference int. w. LK}}~\cite{Kanade81iuw} (CPU)\label{subfig:difference_lukas_cpu_car}]{%
			\includegraphics[width=.15\textwidth]{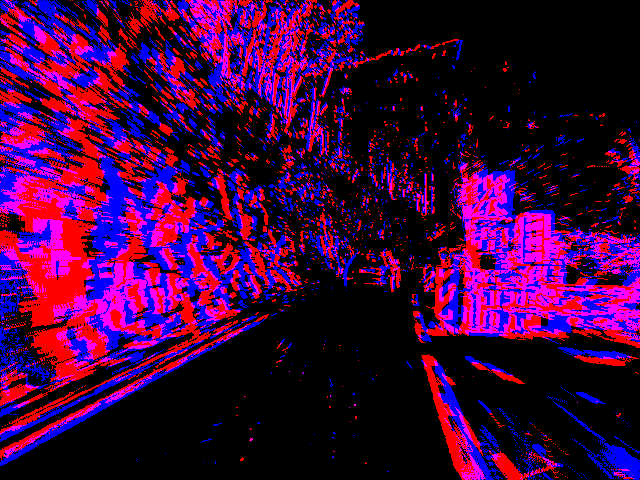}}
		\hfill
		\subfloat[{\color{densecolor}{Dense int. w. Farneback}}~\cite{Farnebck03scia} (GPU)\label{subfig:dense_farneback_gpu_car}]{%
			\includegraphics[width=.15\textwidth]{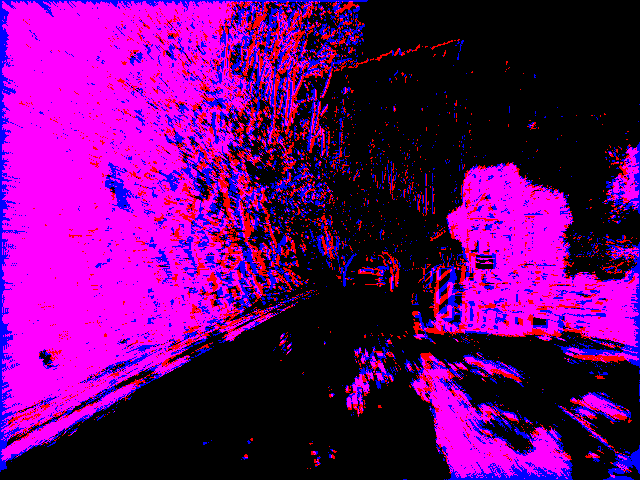}}
		\hfill
		\subfloat[{\color{sparsecolor}{Sparse int. w. LK}}~\cite{Kanade81iuw} (GPU)\label{subfig:sparse_lukas_gpu_car}]{%
			\includegraphics[width=.15\textwidth]{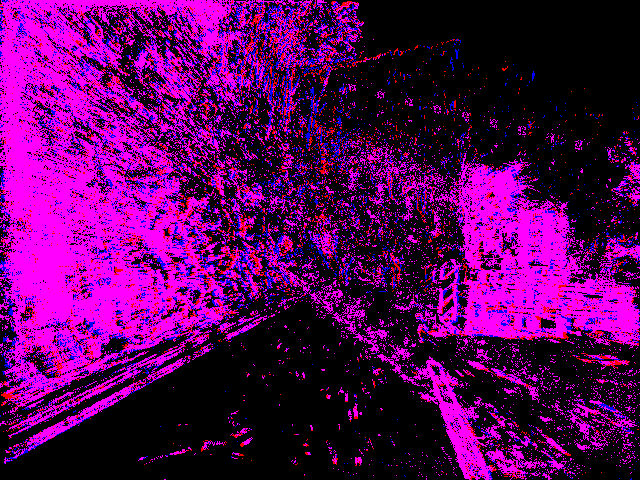}}
		\hfill
		\subfloat[{\color{differencecolor}{Difference int. w. LK}}~\cite{Kanade81iuw} (GPU)\label{subfig:difference_lukas_gpu_car}]{%
			\includegraphics[width=.15\textwidth]{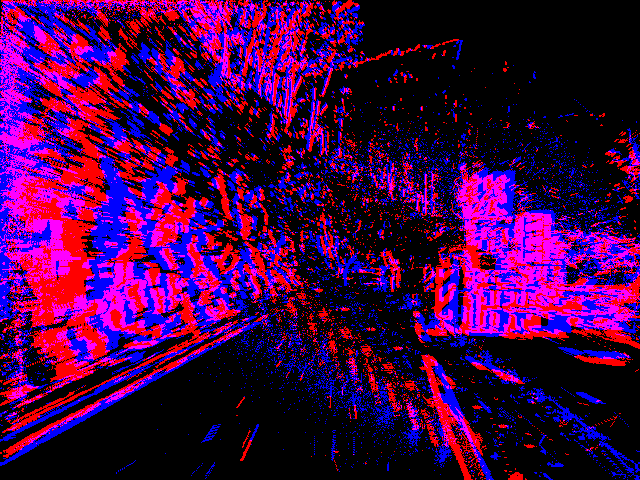}}
		\hfill
		\subfloat[{\color{othercolor}{v2e}}~\cite{Hu21cvprw} (for reference)\label{subfig:v2e_car}]{%
			\includegraphics[width=.15\textwidth]{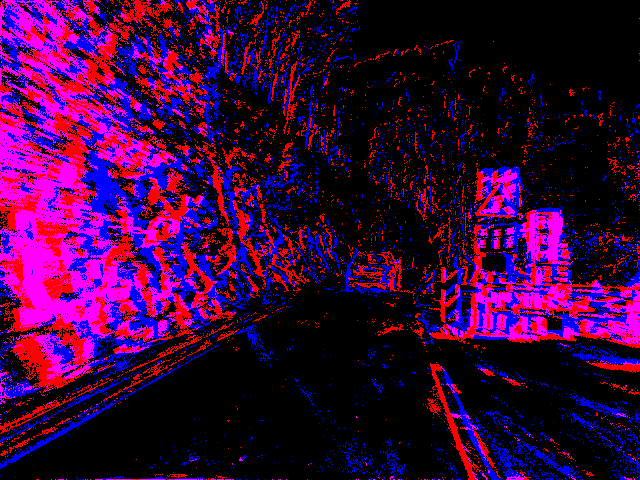}}
		
		\caption{Qualitative comparison of the output of the different simulation methods on a high events-per-frame video sequence depicting a car driving through Interlaken (composed from the DSEC~\cite{Gehrig21ral} dataset).
			LK stands for Lucas Kanade~\cite{Kanade81iuw}.
			The accumulation time was set to mimic the frame rate of the frame camera ($1/\text{fps}$).
			}
		\label{fig:qualitative_comparison_1}
	\end{figure*}
	\begin{figure*}[ht!]
		\centering
		\subfloat[Grayscale frame (for reference)\label{subfig:grayscale_tennis}]{%
			\includegraphics[width=.15\textwidth]{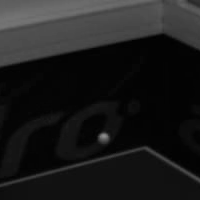}}
		\hfill
		\subfloat[Ground truth accumulated events frame \label{subfig:accum_events_ball}]{%
			\includegraphics[width=.15\textwidth]{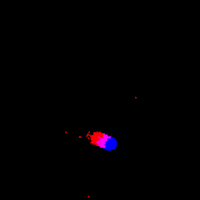}}
		\hfill
		\subfloat[\color{basiccolor}{Difference frame}\label{subfig:difference_frame_tennis}]{%
			\includegraphics[width=.15\textwidth]{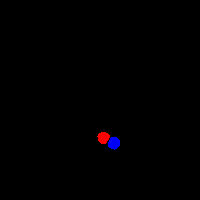}}
		\hfill
		\subfloat[{\color{densecolor}{Dense int. w. DIS}}~\cite{Kroeger16eccv} on lq. preset\label{subfig:dense_dis_low_tennis}]{%
			\includegraphics[width=.15\textwidth]{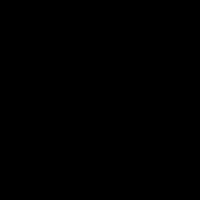}}
		\hfill
		\subfloat[{\color{densecolor}{Dense int. w. DIS}}~\cite{Kroeger16eccv} on hq. preset\label{subfig:dense_dis_high_tennis}]{%
			\includegraphics[width=.15\textwidth]{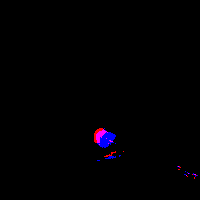}}
		\hfill	
		\subfloat[{\color{densecolor}{Dense int. w. Farneback}}~\cite{Farnebck03scia} (CPU)\label{subfig:dense_farneback_cpu_tennis}]{%
			\includegraphics[width=.15\textwidth]{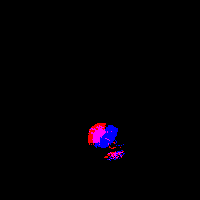}}

		\subfloat[{\color{sparsecolor}{Sparse int. w. LK}}~\cite{Kanade81iuw} (CPU)\label{subfig:sparse_lukas_cpu_tennis}]{%
			\includegraphics[width=.15\textwidth]{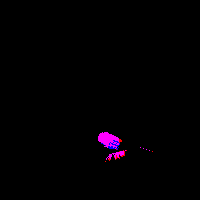}}
		\hfill
		\subfloat[{\color{differencecolor}{Difference int. w. LK}}~\cite{Kanade81iuw} (CPU)\label{subfig:difference_lukas_cpu_tennis}]{%
			\includegraphics[width=.15\textwidth]{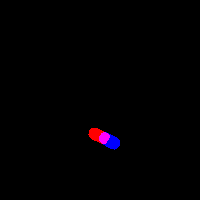}}
		\hfill
		\subfloat[{\color{densecolor}{Dense int. w. Farneback}}~\cite{Farnebck03scia} (GPU)\label{subfig:dense_farneback_gpu_tennis}]{%
			\includegraphics[width=.15\textwidth]{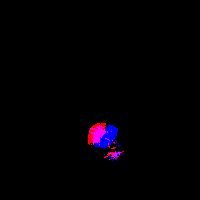}}
		\hfill
		\subfloat[{\color{sparsecolor}{Sparse int. w. LK}}~\cite{Kanade81iuw} (GPU)\label{subfig:sparse_lukas_gpu_tennis}]{%
			\includegraphics[width=.15\textwidth]{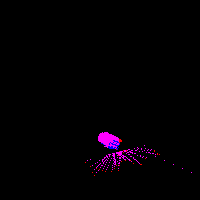}}
		\hfill
		\subfloat[{\color{differencecolor}{Difference int. w. LK}}~\cite{Kanade81iuw} (GPU)\label{subfig:differencee_lukas_gpu_tennis}]{%
			\includegraphics[width=.15\textwidth]{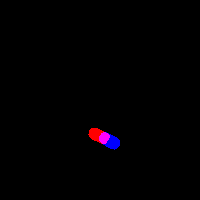}}
		\hfill
		\subfloat[{\color{othercolor}{v2e}}~\cite{Hu21cvprw} (for reference)\label{subfig:v2e_tennis}]{%
			\includegraphics[width=.15\textwidth]{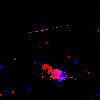}}
		\hfill
		
		\caption{Qualitative comparison of the output of the different simulation methods on a low events-per-frame video sequence depicting a flying table tennis ball.
			LK stands for Lucas Kanade~\cite{Kanade81iuw}.
			The accumulation time was set to mimic the frame rate of the frame camera ($1/\text{fps}$).
			}
		\label{fig:qualitative_comparison_2}
	\end{figure*}
		
	\textbf{Discussion} As can be seen in the results shown in \cref{fig:zurich_times} and \cref{fig:ball_times}, the runtimes differ substantially depending on the event simulation method and their configuration.
	
	Starting with the dynamic dataset, the dense interpolation method with DIS~\cite{Kroeger16eccv} on its lowest quality setting is the fastest configuration for the dense approach, closely followed by the one with Farneback~\cite{Farnebck03scia} (on the GPU).
	DIS takes more than three times as long with the highest quality setting.
	The sparse interpolation method on the GPU is a bit slower than the dense one.
	The sparse interpolation method on the CPU performs the worst of all the techniques with different settings.
	Since there is a lot of movement in this driving car dataset, we expect the sparse interpolation method to be worse.
	As mentioned in \cref{subsec:sparse_interpolation}, sparse optical flow benefits only if the number of pixels that have changed is substantially lower than the total number of pixels.
	In this experiment, this is not the case.
	The GPU and CPU runtimes of the difference interpolation method are better compared to the corresponding runtimes of the sparse interpolation method, with the GPU version outperforming the CPU version.
	
	On the less dynamic dataset, the runtimes for the dense interpolation method are all longer because the resolution of the table tennis video is larger ($1280 \times 720$ pixels compared to $640 \times 480$ pixels).
	For the dense interpolation method, the relative comparisons remain the same.
	As can be seen in \cref{fig:ball_times}, with runtimes above 80ms, the dense optical flow algorithms are too slow to be used in real-time applications in such scenarios.
	However, the dense method with DIS with the high quality preset is still three orders of magnitude faster than vid2e~\cite{Gehrig20cvpr} and v2e~\cite{Hu21cvprw}.
	Since there is only one moving ball in front of a static background in this dataset, the sparse interpolation method only needs to estimate the motion of the pixels representing the ball.
	As expected, the sparse interpolation method performs very well.
	The difference interpolation method, with its strict assumption met on this dataset, achieves the best runtimes.
	
	\subsection{QUALITATIVE COMPARISON}\label{subsec:qualitative_comparison}
	
	For qualitative comparison, the outputs of the different methods are shown in \cref{fig:qualitative_comparison_1} for the dynamic dataset and in \cref{fig:qualitative_comparison_2} for the less dynamic dataset.
	We used the same datasets as in \cref{subsec:speed}.
	The accumulation time was set to mimic the frame rate of the frame camera ($1/\text{fps}$).
	
	\textbf{Discussion} Starting with the dynamic dataset, the output of the dense interpolation methods, shown in \cref{subfig:dense_farneback_cpu_car,subfig:dense_farneback_gpu_car,subfig:dense_dis_low_car,subfig:dense_dis_high_car}, bear a very strong resemblance to the ground truth accumulated event frame, shown in \cref{subfig:accum_events_car}.
	The positive and negative overlapping events, missing in the \textit{difference frame} (\cref{subfig:difference_frame_car}), are primarily generated from the interpolation.
	The impact of the different optical flow algorithms is relatively small, with the main difference being the amount of motion registered in the street area of the frames.
	The output of the sparse interpolation methods, shown in \cref{subfig:sparse_lukas_cpu_car,subfig:sparse_lukas_gpu_car}, also bear a very strong resemblance to the ground truth accumulated event frame, shown in \cref{subfig:accum_events_car}.
	The output is noticeably noisier than the one of the dense interpolation methods, which is to be expected because of the sparse interpolation.
	If neighboring pixels are just under the threshold, they will not be interpolated and therefore do not emit an event.
	Additionally, some artifacts appear when Lucas Kanade~\cite{Kanade81iuw} is run on the GPU.
	These artifacts are probably a side product because only patches of the original image are sent to the GPU by OpenCV.
	The difference interpolation method performs quite poorly on this dataset, as can be seen in \cref{subfig:difference_lukas_cpu_car,subfig:difference_lukas_gpu_car}.
	This is mainly due to the constraints not being met.
	The perspective of the objects in the scene changes quite drastically as the car is driving by them.
	However, some events between the two frames are still emitted, and the output is a clear improvement over the simple \textit{difference frame}.
	The output of v2e comes closest to the one of the difference interpolation method from a visual point of view.
	
	In the less dynamic dataset, the dense interpolation method does not perform as well, shown in \cref{subfig:dense_farneback_cpu_tennis,subfig:dense_farneback_gpu_tennis,subfig:dense_dis_low_tennis,subfig:dense_dis_high_tennis}.
	The dense interpolation method with DIS with the low-quality preset cannot even detect any motion, making the ball completely invisible, as seen in \cref{subfig:dense_dis_low_tennis}.
	With the Farneback optical flow algorithm, the dense interpolation method manages to estimate some motion in the ball but creates an event cloud around it (\cref{subfig:dense_farneback_cpu_tennis,subfig:dense_farneback_gpu_tennis}).
	Only the dense interpolation method with DIS on the high-quality preset produces a favorable result, as seen in \cref{subfig:dense_dis_high_tennis}.
	These inaccuracies can be attributed to the dense optical flow algorithms.
	Traditional dense optical flow algorithms find the frame's general motion in patches.
	This is especially a problem for DIS~\cite{Kroeger16eccv} with the low quality preset as the ball is not even registered because of the large patch size and the fact that the surrounding pixels do not move with the ball.
	The output of the sparse interpolation method has some artifacts due to the interpolation of overlapping subjects.
	One object represents where the ball has moved in the current frame, and one object where the ball was in the previous frame.
	The GPU implementation has some additional artifacts similar to the ones in the dynamic dataset.
	The output of the difference interpolation, shown in \cref{subfig:difference_lukas_cpu_tennis,subfig:differencee_lukas_gpu_tennis}, makes it apparent that this method is well suited for this kind of scene.
	It is the only algorithm able to correctly interpolate the ball's position and create the events that lie between the two frames.
	As seen in \cref{subfig:v2e_tennis}, v2e produces quite some artifacts and does not look as similar to the ground truth compared to most of our methods.

	\subsection{STATISTICAL COMPARISON}\label{subsec:statistical_comparison}
	
	In this section, we present a statistical comparison of our event simulation methods with real events and the output of vid2e~\cite{Gehrig20cvpr} and v2e~\cite{Hu21cvprw}.
	We use the metric of events per pixel per second $(\frac{\text{events}}{\text{pixel} \cdot s})$ introduced in~\cite{Stoffregen20eccv} as comparison for two short sequences.
	Event count is a necessary but not sufficient condition for a faithful simulation of an event camera.
	We took $2s$ ($40$ frames) from the {\it interlaken\_00\_c} sequence from~\cite{Gehrig21ral} for high events-per-frame data and $1.13s$ ($170$ frames) from our flying table tennis ball dataset for low events-per-frame data.
	We adjusted the threshold values $C_{\text{pos}}$ and $C_{\text{neg}}$ for every simulation method to get similar mean values as the ones from the real events.
	We list the mean and standard deviation as well as the used threshold values $C_{\text{pos}}$ and $C_{\text{neg}}$ (unless unknown) in \cref{tab:statistics_car} for the high events-per-frame data and in \cref{tab:statistics_ball} for the low events-per-frame data.
	
	\textbf{Discussion} As can be seen in \cref{tab:statistics_car}, for the high events-per-frame dataset, our simulation methods tend to have a bit of a lower standard deviation compared to the real events, v2e and vid2e.
	On the low events-per-frame dataset, however, most of our simulation methods have standard deviations closer to the one of the real events compared to v2e and vid2e.
	
	\begin{table}[h!]
		\caption{Event statistics on a high events-per-frame video sequence}
		\begin{tabular}{l|ll|ll}
			\cline{2-3}
			& \multicolumn{2}{l|}{\textbf{Events / (pixel $\cdot$ s)}}                &                                     &                                     \\ \hline
			\multicolumn{1}{|l|}{\textbf{Method}}                                & \multicolumn{1}{l|}{\textbf{mean}} & \textbf{std. dev.} & \multicolumn{1}{l|}{\textbf{$C_{\text{pos}}$}} & \multicolumn{1}{l|}{\textbf{$C_{\text{neg}}$}} \\ \hline
			\multicolumn{1}{|l|}{Real events}                                    & \multicolumn{1}{l|}{62.7933}       & 82.7128            & \multicolumn{1}{l|}{?}               & \multicolumn{1}{l|}{?}               \\ \hline
			\multicolumn{1}{|l|}{\color{densecolor}Dense int. w. DIS, lq. preset}       & \multicolumn{1}{l|}{56.3079}       & 58.9868            & \multicolumn{1}{l|}{3}              & \multicolumn{1}{l|}{-3}             \\ \hline
			\multicolumn{1}{|l|}{\color{densecolor}Dense int. w. DIS, hq. preset}       & \multicolumn{1}{l|}{65.6304}       & 66.793             & \multicolumn{1}{l|}{3}              & \multicolumn{1}{l|}{-3}             \\ \hline
			\multicolumn{1}{|l|}{\color{densecolor}Dense int. w. Farneback}         & \multicolumn{1}{l|}{65.999}        & 67.4279            & \multicolumn{1}{l|}{3}              & \multicolumn{1}{l|}{-3}             \\ \hline 
			\multicolumn{1}{|l|}{\color{sparsecolor}Sparse int. w. LK}     & \multicolumn{1}{l|}{59.3083}       & 52.3827            & \multicolumn{1}{l|}{6}              & \multicolumn{1}{l|}{-6}             \\ \hline 
			\multicolumn{1}{|l|}{\color{differencecolor}Difference int. w. LK} & \multicolumn{1}{l|}{59.3034}       & 55.0075            & \multicolumn{1}{l|}{19}             & \multicolumn{1}{l|}{-19}            \\ \hline 
			\multicolumn{1}{|l|}{{\color{othercolor}v2e}~\cite{Hu21cvprw}}                                            & \multicolumn{1}{l|}{58.8313}       & 78.3513            & \multicolumn{1}{l|}{0.1}            & \multicolumn{1}{l|}{0.1}            \\ \hline
			\multicolumn{1}{|l|}{{\color{othercolor}vid2e}~\cite{Gehrig20cvpr}}                                          & \multicolumn{1}{l|}{63.6588}       & 77.9866            & \multicolumn{1}{l|}{0.14}           & \multicolumn{1}{l|}{0.14}           \\ \hline
		\end{tabular}
		\label{tab:statistics_car}
	\end{table}
	
	\begin{table}[h!]
		\caption{Event statistics on a low events-per-frame video sequence}
		\begin{tabular}{l|ll|ll}
			\cline{2-3}
			& \multicolumn{2}{l|}{\textbf{Events / (pixel $\cdot$ s)}}                &                                     &                                     \\ \hline
			\multicolumn{1}{|l|}{\textbf{Method}}                                & \multicolumn{1}{l|}{\textbf{mean}} & \textbf{std. dev.} & \multicolumn{1}{l|}{\textbf{$C_{\text{pos}}$}} & \multicolumn{1}{l|}{\textbf{$C_{\text{neg}}$}} \\ \hline \hline
			\multicolumn{1}{|l|}{Real events}                                    & \multicolumn{1}{l|}{0.046}         & 1.0228             & \multicolumn{1}{l|}{43}             & \multicolumn{1}{l|}{-17}            \\ \hline
			\multicolumn{1}{|l|}{\color{densecolor}Dense int. w. DIS, lq. preset}       & \multicolumn{1}{l|}{0.0472}        & 1.165              & \multicolumn{1}{l|}{13}             & \multicolumn{1}{l|}{-13}            \\ \hline
			\multicolumn{1}{|l|}{\color{densecolor}Dense int. w. DIS, hq. preset}       & \multicolumn{1}{l|}{0.0462}        & 0.8783             & \multicolumn{1}{l|}{35}             & \multicolumn{1}{l|}{-35}            \\ \hline
			\multicolumn{1}{|l|}{\color{densecolor}Dense int. w. Farneback}         & \multicolumn{1}{l|}{0.0445}        & 0.574              & \multicolumn{1}{l|}{19}             & \multicolumn{1}{l|}{-19}            \\ \hline 
			\multicolumn{1}{|l|}{\color{sparsecolor}Sparse int. w. LK}     & \multicolumn{1}{l|}{0.0467}        & 0.439              & \multicolumn{1}{l|}{27}             & \multicolumn{1}{l|}{-27}            \\ \hline 
			\multicolumn{1}{|l|}{\color{differencecolor}Difference int. w. LK} & \multicolumn{1}{l|}{0.0461}        & 0.8785             & \multicolumn{1}{l|}{112}            & \multicolumn{1}{l|}{-112}           \\ \hline 
			\multicolumn{1}{|l|}{{\color{othercolor}v2e}~\cite{Hu21cvprw}}                                            & \multicolumn{1}{l|}{0.0455}        & 0.4912             & \multicolumn{1}{l|}{0.575}          & \multicolumn{1}{l|}{0.575}          \\ \hline
			\multicolumn{1}{|l|}{{\color{othercolor}vid2e}~\cite{Gehrig20cvpr}}                                          & \multicolumn{1}{l|}{0.0318}        & 1.776              & \multicolumn{1}{l|}{5.7}            & \multicolumn{1}{l|}{-5.7}           \\ \hline
		\end{tabular}
		\label{tab:statistics_ball}
	\end{table}
	
	\section{CONCLUSION}\label{sec:conculsion}
	
	This work presented multiple event simulator methods that can run in under $30$ms. 
	Compared to existing event simulators, ours can be used in real-time robotics applications while remaining competitive in the quality assessment.
	We have shown the influence of the resolution and the dynamics of the scene on the runtime of the simulators, the quality of the simulated events, and the statistics of the events.
	
	Depending on the camera setup and the dynamics of the scene, we have the following suggestions:
	For scenes with high dynamics, e.g., an event camera mounted on a driving car, we recommend the dense interpolation method with Farneback on the GPU.
	In scenes with fewer dynamics, e.g., tracking an object with a static event camera, we recommend using the difference interpolation method.
	For use cases in between, some experiments might be needed to find the sweet spot between a fast runtime and a high quality of the simulated events.
	The sparse interpolation method might be a good choice.
	
	With this work, we have made a first step towards event simulators for real-time robotics applications.
	We hope our work enables others to develop new algorithms for event-based data in real-time use cases.

	


	
	
	


	\bibliographystyle{IEEEtran}
	\bibliography{bibliography}
	
\end{document}